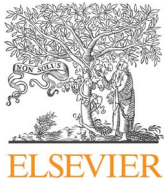
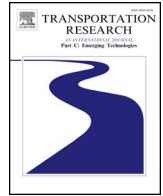

# Inferring transportation modes from GPS trajectories using a convolutional neural network

Sina Dabiri[*], Kevin Heaslip

*Charles E. Via, Jr. Department of Civil and Environmental Engineering, 301, Patton Hall, Virginia Tech, Blacksburg, VA 24061, United States*

ABSTRACT

Identifying the distribution of users' transportation modes is an essential part of travel demand analysis and transportation planning. With the advent of ubiquitous GPS-enabled devices (e.g., a smartphone), a cost-effective approach for inferring commuters' mobility mode(s) is to leverage their GPS trajectories. A majority of studies have proposed mode inference models based on hand-crafted features and traditional machine learning algorithms. However, manual features engender some major drawbacks including vulnerability to traffic and environmental conditions as well as possessing human's bias in creating efficient features. One way to overcome these issues is by utilizing Convolutional Neural Network (CNN) schemes that are capable of automatically driving high-level features from the raw input. Accordingly, in this paper, we take advantage of CNN architectures so as to predict travel modes based on only raw GPS trajectories, where the modes are labeled as walk, bike, bus, driving, and train. Our key contribution is designing the layout of the CNN's input layer in such a way that not only is adaptable with the CNN schemes but represents fundamental motion characteristics of a moving object including speed, acceleration, jerk, and bearing rate. Furthermore, we ameliorate the quality of GPS logs through several data preprocessing steps. Using the clean input layer, a variety of CNN configurations are evaluated to achieve the best CNN architecture. The highest accuracy of 84.8% has been achieved through the ensemble of the best CNN configuration. In this research, we contrast our methodology with traditional machine learning algorithms as well as the seminal and most related studies to demonstrate the superiority of our framework.

## 1. Introduction

Travel mode choice is one of the principal traveler's behavior attributes in travel demand analysis, transport planning, and traffic management. By inferring the travel mode distribution, transportation agencies are able to generate appropriate strategies to alleviate users' travel time, traffic congestion, and air pollution. For example, a clear benefit of travel mode analysis is to identify regions with high auto dependency and encourage public transport ridership by improving transit systems (Eluru et al., 2012). Furthermore, suitable policies such as HOV lanes can be taken during the peak-period congestion according to existing mode shares. The knowledge of the transport mode choice has traditionally been obtained through household surveys or phone interviews. However, interviewing households is a time-consuming and expensive method that usually results in a low response rate and incomplete information, which calls for a cost-effective technology such as Global Positioning System (GPS) that is able to collect travel data while reducing labor and time costs.

GPS is a ubiquitous positioning tool that records spatiotemporal information of moving objects carrying a GPS-enabled device

(e.g., a smartphone). The main advantageous of smart phones, compared to other GPS-equipped devices, is its enormous market penetration rate in a large number of countries and being relatively close to users nearly all of the time. As a consequence, such a dominant and area-wide sensing technology is capable of creating massive trajectory data of vehicles and people. A GPS trajectory, also called movement, of an object is constructed by connecting GPS points of their GPS-enabled device. A GPS point, here, is denoted as (*lat, long, t*), where *lat, long,* and *t* are latitude, longitude, and timestamp, respectively. The study of individuals' mobility patterns from GPS datasets has led to a variety of behavioral applications including learning significant locations, anomaly detection, location-based activity recognition, and identification of transport modes (Lin and Hsu, 2014), in which the latter is the focus of this study. Nonetheless, GPS devices can only record time and positional characteristics of travels without any explicit information on utilized transport modes. This necessitates employing data processing and mining algorithms to extract hidden knowledge about transport modes from raw GPS data.

Many of proposed inference models on detecting transport modes by means of only GPS sensors include two steps. In the first step, a pool of attributes (e.g., velocity, acceleration, heading change rate, and stop rate) (Zheng et al., 2008a,b), are computed from GPS logs. In the second step, the extracted features are fed into a learning algorithm to estimate the transportation mode. Unlike many classification problems that a majority of features have already been computed and included in the dataset, raw GPS trajectories contain only a series of chronologically ordered points without any explicit features such as speed and acceleration. This fact has required researchers to manually identify and formulate a set of features before using a machine learning technique for the classification task. However, the hand-crafted features may not necessarily distinguish between various transportation modes since they are vulnerable to traffic and environmental conditions (Zheng et al., 2008b). Considering a congested traffic condition, for instance, the maximum velocity of a car might be equal to the bicycle and walk modes. To address this issue, one solution is to extract more features from a GPS track (e.g., top five velocities rather than a single maximum velocity). However, if a lot of features are generated to cover more aspects of GPS trajectories, the challenge of applying an effective dimensionality reduction process needs to be met. Manual features are typically produced based on feature engineering, which is a concept upon biased engineering justification and commonsense knowledge of the real world for creating features and making patterns more visible for learning algorithms. Since the ultimate performance of machine learning algorithms is contingent on how much accurate the hand-crafted features are, the effectiveness proof of such features is not trivial. One way to address the above-mentioned issues is to exploit deep learning algorithms that are able to automatically and without any human interference extract multiple levels of data representations.

Indeed, the input layer (i.e., input features) in deep learning architectures is the raw object (e.g., images) rather than a set of hand-crafted features. The key role of deep learning techniques is to encode object's raw and low-level features (e.g., raw image pixels) to multiple levels of efficient and high-level features. This is the most salient attribute of deep learning algorithms that distinguishes them from classical machine learning algorithms. Thus, new representations of the raw data, which are more effective for the classification task, are generated by machines instead of humans. It should be noted that the learned features in the last layer play the same role as hand-crafted features, which are fed into activation functions (e.g., SVM or softmax) to compute class scores.

In this paper, we propose a Convolutional Neural Network (CNN) architecture to predict the transportation mode(s) used in an individual's trip from their raw GPS trajectories, in which modes are categorized into walk, bike, bus, driving, and train. The CNN is a type of deep learning techniques that has achieved great success in the fields of computer vision (Krizhevsky et al., 2012) and natural language processing (Kim, 2014). We envision to investigate the capability of CNNs in representation learning and transport mode classification of raw GPS data. Unlike other fields (e.g., image classification) that images are easily utilized as the input layer, the key challenge in this study is to structure raw GPS tracks into a format that is not only acceptable for CNN architectures but efficient enough to represent fundamental motion characteristics of a moving object. In our methodology, an instance comprises four channels of kinematic features including speed, acceleration, jerk, and bearing rate. Stacking these channels yields a standard arrangement for the CNN scheme that also describes people's motion characteristics. After pre-processing data and designing a suitable layout for each instance, we come up with an effective CNN architecture so as to attain state-of-the-art accuracy on the GPS dataset collected by the Microsoft GeoLife project (Zheng et al., 2008b).

The rest of this article is organized as follows. After reviewing related works in Section 2, we set out details of our framework in Section 3, including preparing the input layer, applying data pre-processing steps, explaining settings of CNN layers, and generating several CNN configurations for our application. In Section 4, we evaluate our proposed CNN architecture on the GeoLife trajectory dataset. Comparing our results with classical machine learning algorithms and previous research is also carried out in Section 4. Finally, we conclude the paper in Section 5.

## 2. Literature review

A large and growing body of literature has proposed numerous frameworks for inferring the commuters' transport mode based on various data sources including raw GPS trajectories (Zheng et al., 2008a; Endo et al., 2016; Xiao et al., 2017), mobile phone's accelerometers (Nick et al., 2010), and mobile phone's GSM data (Sohn et al., 2006). For performing the mode classification task, a wide range of traditional supervised mining algorithms have been applied, including rule-based methods, fuzzy logic, decision tree, Bayesian belief network, multilayer perceptron, and support vector machine (Wu et al., 2016). Furthermore, some of the mobility detection research has integrated multiple sources to ameliorate the classification quality. For example, Stenneth et al. (2011) exploited the GPS and GIS information for building their mode detection scheme while (Feng and Timmermans, 2013) combined GPS and accelerometer data to generate a model that outperforms GPS-only information. In addition to the GPS and accelerometer, the data from other mobile phone sensors such as gyroscope, rotation vector, and magnetometer have been deployed in distinguishing between different transportation modes (Jahangiri and Rakha, 2014; Eftekhari and Ghatee, 2016). Integrating such spatial and temporal information with socio-demographic characteristics of travelers can also lead to generating richer travel mode detection models (Bantis and Haworth, 2017). However, the methodologies using only one type of sensor can be more practical since accessing to several data sources may not be possible in many cities (Xiao et al., 2017). As this study seeks to design a mobility inference model using only raw GPS data, we only review the studies that examine the capability of the GPS sensor for distinguishing between users' transportation modes. A comprehensive and systematic review of existing techniques of travel mode recognition based on GPS data is available in the reference (Wu et al., 2016). The paper provides an excellent comparison of various approaches in three categories including GPS data preprocessing, trip/segmentation identification, and travel mode detection.

Zheng et al. (2008b) proposed a solid framework to automatically infer transportation mode(s) from users' GPS trajectories. In their seminal study, first, a change-point-based segmentation algorithm is applied to divide a trip into segments with distinct transportation modes. Afterward, features of each segment are extracted, including the mean and variance of the velocity, the expectation of velocity, the top three velocities and accelerations. Considering each segment with its features and its assigned transportation label as an independent instance, various traditional classification algorithms (e.g., decision tree) were deployed to train the mode inference model. In a follow up study, they improved the performance accuracy of their mode inference framework by involving more robust trajectory's attributes as well as applying a graph-based post-processing algorithm (Zheng et al., 2008a). The new features are composed of the rate of change in the heading direction, the stop rate, and the velocity change rate. In the post-processing algorithm, a graph is built based on clustering users' change points into nodes and assigning the probability distribution of different modes on edges. Using the GPS data, Sun and Ban (2013) utilized the features related to only accelerations and decelerations (e.g., the proportions of accelerations and decelerations larger than 1 meter per square second, and the standard deviations of accelerations and decelerations) to classify vehicles into general trucks and passenger cars. A recent study by Xiao et al. (2017) demonstrated that tree-based ensemble classification algorithms outperform traditional ones such as the decision tree. Their major contribution was to augment the number of GPS trajectory features to 111 using statistical methods. The features were categorized into global attributes, extracted from descriptive statistics for the entire trajectory, and local features, extracted from the profile decomposition that describes the movement behavior. Mäenpää et al. (2017) sought to select the most significant features from three sets of potential features extracted from GPS trajectories. According to the statistical tests, frequency-domain features were found to be significant in classifying transportation modes while auto- and cross-correlations, kurtoses and skewnesses of speed and acceleration were not useful.

A few recent studies have attempted to integrate the hand-crafted features developed by Zheng et al. (2008a) with high-level features extracted through deep learning architectures (Endo et al., 2016; Hao et al., 2017). Endo et al. (2016) exploited a fully-connected Deep Neural Network (DNN) to automatically extract high-level features. Their core idea was to convert a raw GPS trajectory into a 2-D image structure as the input for the DNN model. The pixel value in the created trajectory image is equivalent to the duration time that a user stays in the location of the pixel. After integrating the image-based deep features with the manual features, a traditional classifier is deployed to predict transportation labels. However, the method of creating trajectory images suffers from some drawbacks. The pixel values contain only the duration time without any motion attributes and a theoretical concept. In other words, the technique fails to take spatiotemporal information into account that can be measured by manipulation of latitude, longitude, and timestamp. Lack of any motion information (e.g., speed and acceleration) in the input layer leads to generating irrelative features to the mission of mode inference. Moreover, in order to prevent having biased pixel values, only GPS points with a fixed time interval between two consecutive GPS points are sampled, which results in losing valuable information.

In a similar study conducted by Hao et al. (2017), deep features are obtained by transforming point-level features (PF) with the aid of the sparse auto-encoder. They defined PF as a time series of speed, head change, time interval, and distance of the GPS points. The deep features are then aggregated using a simple CNN before being combined with the hand-crafted features. Finally, both types of features are fed into a DNN to infer the transportation mode. Although their work involved the motion characteristics in the original input features, multicollinearity arises from a high and linear correlation between speed, time interval, and distance. This issue degrades the quality of the feature learning in auto-encoder and CNN techniques. Furthermore, the distance and time interval obtained by means of GPS-equipped devices are not discernable features for detecting transport modes. For instance, a car moving in a crowded area with huge skyscrapers may record two consecutive GPS points with a large distance while a walking individual can record a short distance in the environment with no signal blockages. Note that the distance of two successive GPS points for the car is shorter than the walk mode in a normal situation. It is worth noticing that the highest accuracy of both deep-learning-based studies (Endo et al., 2016; Hao et al., 2017) are still lower than the work of Zheng et al. (2008b), which is purely based on manual features.

To tackle the above-mentioned shortcomings, we propose an efficient CNN architecture with various types of layers that receive an informative input layer with kinematic characteristics. Both feature learning and classification task are executed in an integrated CNN architecture. To show the superiority of CNNs, we also compare the predication quality of our CNN scheme with traditional supervised learning algorithms that have broadly been used in transportation-mode-inference models.

## 3. Methodology

### 3.1. Preparing the input samples and applying data processing

The raw GPS data for each user consists of chronologically ordered points that have been collected over a time period. First, the user's GPS track is divided into trips if the time interval between two consecutive GPS points exceeds a pre-defined threshold. Each trip is divided into segments based on the change in the transportation mode (i.e., the GPS series of each segment contains only one transportation label). In the CNN architectures, all samples are required to have the same size. Considering a constant length for all instances, each segment is either sub-divided into a pre-defined number of GPS points or padded with zero values to have the same size as other segments. After generating segments with the same size, we compute the motion characteristics of each GPS point using the tuple (*lat*, *long*, *t*) of two consecutive points.

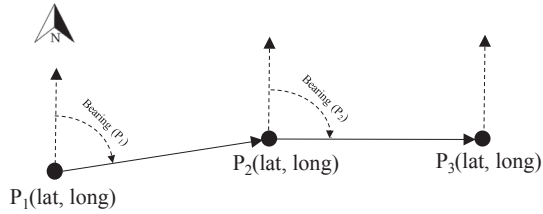

**Fig. 1.** Bearing between two consecutive GPS points.

We utilize the Vincenty's formula (Vincenty, 1975) for calculating the geographical distance between two succeeding GPS points $P_1$ and $P_2$. Denoting the time difference between $P_1$ and $P_2$ as $\Delta t$, the first three motion features of $P_1$ are then calculated based on the following equations:

$$S_{P_1} = \frac{Vincenty(P_1,P_2)}{\Delta t} \tag{1}$$

$$A_{P_1} = \frac{S_{P_2}-S_{P_1}}{\Delta t} \tag{2}$$

$$J_{P_1} = \frac{A_{P_2}-A_{P_1}}{\Delta t} \tag{3}$$

where $S_p$, $A_p$, and $J_p$ represent the speed, acceleration/deceleration, and jerk of the point *P*, respectively. Jerk, the rate of change in the acceleration, is a significant factor in safety issues such as critical driver maneuvers and passengers' balance in public transportation vehicles (Bagdadi and Várhelyi, 2013). Accordingly, we engage it in our calculation as a distinguishable attribute among various modes.

The rate of change in the heading direction of different transportation modes varies. For example, cars and buses have to move only alongside existing streets while people with walk or bike modes alter their directions more frequently (Zheng et al., 2008a). To quantify this disparity among modes, we introduce the bearing rate as the fourth motion attribute. As depicted in Fig. 1, bearing measures the angle between the line connecting two successive points and a reference (e.g., the magnetic or true north). The bearing rate is the absolute difference between the bearings of two consecutive points, which are calculated as follows:

$$y = sine[p_2(long)-p_1(long)]*cosine[p_2(lat)] \tag{4}$$

$$x = cosine[p_1(lat)]*sine[p_2(lat)]-sine[p_1(lat)]*cosine[p_2(lat)]*cosine[p_2(long)-p_1(long)] \tag{5}$$

$$Bearing_{(P_1)} = arctan(y,x) \tag{6}$$

$$BR_{(P_1)} = |Bearing_{(P_2)}-Bearing_{(P_1)}| \tag{7}$$

where *sine*, *cosine*, and *arctangent* are trigonometric functions. Latitude (*lat*) and longitude (*long*) needs to be in radians before passing to Eqs. (4)–(6). The output of Eq. (6) is then converted to degree. According to Eqs. (4)–(7) and Fig. 1, the bearing rate (*BR*) is calculated using the information of three consecutive GPS points. Analogously, the above-mentioned formulas are utilized to calculate the motion features of other GPS points in a segment.

Before designing an adaptable input layer for CNN architectures, we need to identify and remove outliers and random errors that have been generated due to several error sources such as satellite or receiver clocks, atmospheric disturbances, and multipath signal reflection. First, we apply the following data preprocessing steps to detect and remove invalid and inaccurate GPS points:

- The GPS data points with the timestamp greater than their next GPS point are identified and discarded.
- Considering a maximum threshold speed for each transport mode, provided in Table 2, the invalid GPS points with an unrealistic large speed are discarded.
- Considering a maximum threshold acceleration for each transport mode, provided in Table 2, the invalid GPS points with an unrealistic large acceleration are discarded.
- After removing the unrealistic GPS points, the segments with the number of GPS points less than a threshold are identified and discarded.

In the second step of data pre-processing, we apply a smoothing kernel to GPS trajectories so as to remove random errors. Smoothing is a process that data points are averaged by their neighbors using a specific kernel function, in which the shape of kernel distinguishes various smoothing techniques. Since we have no sense on how the sequence of motion characteristics for a segment might fluctuate, we need to utilize a kernel that has no pre-defined shape. In this paper, we implement the Savitzky-Golay filter to all segments' sequences. For each data point of a sequence, the method fits a polynomial to a set of input samples laid over an odd-sized window centered at the subject point (Schafer, 2011). Then, the new value of the subject point is computed by evaluating the resulting polynomial at the subject point. The main advantageous of Savitzky-Golay filter is its capability to maintain the original shape and pattern of the signal.

After obtaining the clean and validated segments with high-quality GPS points, we must stack the vectors of the motion features related to each segment so as to create independent samples. Accordingly, a four-channel structure is built for each segment, where each channel represents speed, acceleration/deceleration, jerk, and bearing rate. Thus far, we have achieved to prepare input samples for feeding into CNN architectures. In the next section, we describe our designated CNN configurations that are primarily inspired by Simonyan and Zisserman (2014) and Krizhevsky et al. (2012).

### 3.2. CNN architecture

Although the key idea in the Convolutional Neural Network (CNN) is similar to the ordinary feed-forward artificial neural network, they differ in terms of connectivity patterns between the neurons in adjacent layers. Unlike the traditional multilayer perceptron (MLP) that each node is fully connected to nodes in the previous layer, the CNN takes the advantage of the spatially local correlation by connecting neurons to only a small region of the preceding layer, also called the receptive field. Such a local connectivity between nodes leads to having the smaller number of weights, which mitigates the curse of dimensionality and the overfitting problem.

Typically, a CNN architecture constitutes a sequence of layers in which each layer transforms an input volume to an output volume of neurons using a set of operations. Although various types of layers have been introduced in literature, we describe only those layers that have been used in building our CNN structure, including the input layer, convolutional layer, pooling layer, fully-connected layer, and dropout layer.

#### 3.2.1. Input layer

One distinguishing feature in the CNN is the capability of accepting the input volume in three dimensions: height, width, and depth (channels). The input layer comprises a set of independent samples, where each sample is a GPS segment with four channels including speed, acceleration, jerk, and bearing rate. Each channel has the shape of $(1 \times M)$, where $M$ determines the segment length (i.e., the number of GPS points that forms the segment). Thus, the input shape is a tuple of $(1 \times M \times 4)$. Since the GPS segments contain a different number of GPS points, the value of $M$ may vary over the samples. On the other hand, in the CNN, the input shape for all instances must be the same. For addressing this challenge, all segments are restricted to a fixed size of $M$. Thus, long segments are truncated to $M$ while the shorter ones are padded with zero values. Fig. 2 illustrates the four-channel structure for a GPS segment.

#### 3.2.2. Convolutional layer

The convolutional layer comprises a set of learnable filters. Each neuron in the convolutional layer's output is connected to the small region (receptive field) of the previous layer, where the size of the receptive field is equivalent to the filter. The output value of the neuron is computed by operating dot product between the parameters of the filter and the entries of its receptive field. Convolving the same filter across the whole surface of the input volume creates a 2-dimensional map in the output volume, also called feature map or activation map. Performing similar operations for all layer's filters creates several activation maps. Stacking the feature maps of all filters along the depth dimension creates the 3-d output volume of the layer. Since the size of the input volume is small in our application, we use a small $(1 \times 3 \times C)$ convolution filters throughout for all convolutional layers. For each layer, $C$ indicates the number of channels in the layer's input volume. Using smaller receptive fields leads to reducing the number of parameters and mitigating the overfitting problem (Simonyan and Zisserman, 2014).

The 3-d output shape of each convolutional layer is controlled by using three hyperparameters, which needs to be specified by the user. The depth of the output volume, denoted as $D$, is the first parameter that corresponds to the number of filters used for the convolutional operation. Stride, denoted as *S*, is the number of elements by which the filter is moved at a time over the input volume. The zero-padding, denoted as *P*, is the processing of adding zero values at the start and end of the input volume matrix so as to control the spatial size of the output. In all convolutional layers, we set $S = 1$ and set *P* so that the spatial dimensions of the layer's input do not alter (i.e., the layer's input and output volumes have the same size). But, the number of filters are tuned through a manual search over a variety of CNN configurations as described in Section 3.2.7.

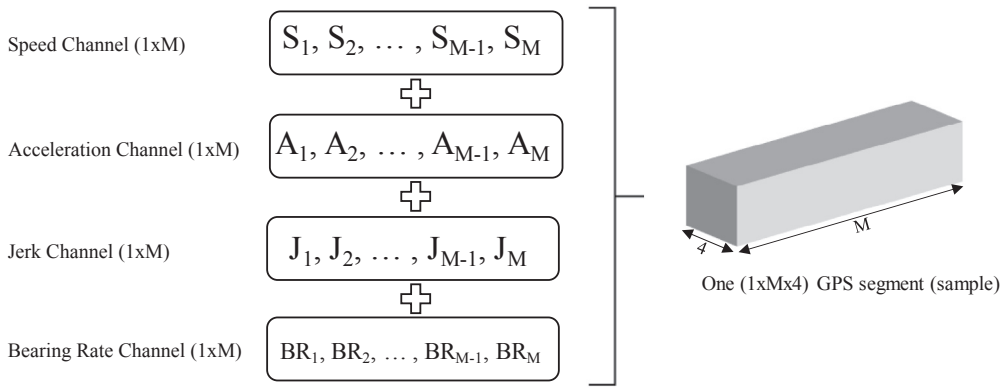

Fig. 2. The four-channel structure for a GPS segment (i.e., an independent sample).

### 3.2.3. Activation layer

After obtaining the convolved output volume, the neurons are often followed by an activation operation with the purpose of introducing the nonlinearity in the model. Among several types of activation function, we utilize the non-saturating nonlinearity $f(x) = \max(0, x)$ in all convolutional layers, where *x* denotes the convolved neurons. The function, which is referred to the Rectified Linear Units (ReLU), replaces all negative values in the feature map by zero. Compared to other functions such as $tanh(x)$ and sigmoid function, the learning rate of the CNN with ReLU is much faster (Krizhevsky et al., 2012).

### 3.2.4. Pooling layer

The objective of pooling layer is to achieve spatial and scale invariance, lower computation, and controlling overfitting by decreasing the dimensionality of each feature map and down sampling the convolution layer spatially (Scherer et al., 2010). Max pooling is the most common kind of spatial pooling that partitions each depth slice of the input volume into non-overlapping vectors, then take the maximum value in each sub-vector. The filter size of the max-pooling layer determines the length of sub-vectors (i.e., the number of entries that the max should be taken over). Hence, the depth of input and output volumes in the pooling layer are the same. In our problem, we insert the max-pooling layer with the filter size (1 × 2) and $S = 1$ periodically between the successive convolutional layers.

### 3.2.5. Fully-connected layer

The CNN architecture can be brought to an end with several Fully-Connected (FC) layers. Like the traditional multilayer perceptron, every neuron in the FC layer is connected to all neurons in the previous layer and computed by the element-wise multiplication. Except for the last FC layer, the rest play the role of extracting features. Subsequently, the extracted high-level features from previous layers are fed into the last FC layer for performing the classification task, in which the softmax activation function is utilized to generate a probability distribution over the transportation labels. Except for the last FC layer, it is not necessary to have the number of the output neurons in FC layers the same as the number of labels.

### 3.2.6. Regularization

Regularization is a process used in an attempt to prevent the overfitting problem in statistical models by explicitly controlling the model complexity and constraining the fitting procedure (Friedman, 2001). Overfitting is a major problem in the CNNs due to a large number of weights and complicated relationships between inputs and outputs. Dropout is the most practical and widely used approach to overcome the overfitting problem in CNNs (Srivastava et al., 2014). The technique drops out the input units, with all its incoming and outgoing connections, with a probability of *P* from the network at each update during the training process. Only the parameters associated to the reduced network is trained at each stage.

Increasing the number of samples is another approach to prevent the overfitting problem. The idea of subdividing the segments into samples with the fixed length *M* is also augmenting the samples more than four times compared to the previous studies (Endo et al., 2016; Zheng et al., 2008a; Xiao et al., 2017). Since the CNN typically work well in large-scale datasets, the data augmentation is an effective strategy due to lack of a massive GPS data set with transportation labels.

Furthermore, we implement the idea behind the bootstrap aggregating algorithm that combines the output of multiple base learners (Breiman, 1996). In our application, the base learner is the best CNN configurations. Thus, the base learners are first trained independently on randomly selected training instances with replacement, and then their softmax class probabilities are averaged to generate the transportation label posteriors.

### 3.2.7. CNN configurations

Setting the size of filters and stride the same as what described above for all layers, a variety of CNN configurations can be built based on the number of layers, the order of layers, and the number of filters in each convolutional layer. We design a broad spectrum of networks to achieve the best one that fits well to our application. Yet, instead of applying an exhaustive and computationally expensive search over all CNN's hyperparameters for identifying their optimal values, we pursue an efficient manual search inspired by one of the most cited papers in the CNN field (Simonyan and Zisserman, 2014). We start with shallow networks and low number of filters and gradually increase the number of layers and filters in each layer to evaluate if the CNN model's beats higher accuracy. For the sake of simplicity, as illustrated in Table 1, we bring only the important configurations out of many tested networks to highlight the effect of four hyperparameters, including the layers' pattern, the absence or presence of a layer, the CNN's depth, and the layers' number of filters, on the prediction quality. Initially, we set the *P* values in dropout layers as 0.5. Yet the final optimal values of *P* in dropout layers are optimized using the grid search method. The overall layout stacks two convolutional layers, following with or without max-pooling and dropout layers, and may repeat this pattern for a few times to increase the CNN's depth. The configurations are completed by adding one to three FC layers. Excluding the last FC layer, outputs of all convolutional and FC layers are activated by ReLU. Apart from the last FC layer, the number of neurons in an FC layer are equal to one-fourth of neurons in its previous flattened layer.

**Table 1**

Multiple CNN configurations per column. The number in the convolutional layers is the number of filters. Filter sizes for all convolutional layers and max-pooling layers are (1 × 3) and (1 × 2), respectively. Number of neurons in FC layers is one-fourth of neurons in its previous flattened layer. Number of neurons in the last FC layer is 5, equal to the number classes. Yes and No indicate the existence or non-existence of a layer. Fraction rate of dropped units in the FC layers is 0.5. Bold elements show the change(s) compared to the previous configuration.

| Input layer | A | B | C | D | E | F | G | H | I |
|---|---|---|---|---|---|---|---|---|---|
| | A pool of (1 × 200 × 4) GPS segments | | | | | | | | |
| Convolutional | 32 | 32 | 32 | 32 | 32 | 32 | 32 | 32 | 32 |
| Convolutional | 32 | 32 | 32 | 32 | 32 | 32 | 32 | 32 | 32 |
| Max-pooling | No | No | No | No | **Yes** | Yes | Yes | Yes | Yes |
| Dropout | No | No | No | No | No | **Yes** | **No** | No | No |
| Convolutional | No | **64** | 64 | 64 | 64 | 64 | 64 | 64 | 64 |
| Convolutional | No | **64** | 64 | 64 | 64 | 64 | 64 | 64 | 64 |
| Max-pooling | No | No | No | No | **Yes** | Yes | Yes | Yes | Yes |
| Dropout | No | No | No | No | No | **Yes** | **No** | No | No |
| Convolutional | No | No | **128** | 128 | 128 | 128 | 128 | 128 | 128 |
| Convolutional | No | No | **128** | 128 | 128 | 128 | 128 | 128 | 128 |
| Max-pooling | No | No | No | No | **Yes** | Yes | Yes | Yes | Yes |
| Dropout | No | No | No | No | No | **Yes** | Yes | Yes | Yes |
| Convolutional | No | No | No | No | No | No | No | **256** | **No** |
| Convolutional | No | No | No | No | No | No | No | **256** | **No** |
| Max-pooling | No | No | No | No | No | No | No | No | No |
| Dropout | No | No | No | No | No | No | No | No | No |
| FC | No | No | No | No | No | No | No | No | **Yes** |
| Dropout | No | No | No | No | No | No | No | No | **Yes** |
| FC | No | No | No | **Yes** | Yes | Yes | Yes | Yes | Yes |
| Dropout | No | No | No | No | No | **Yes** | Yes | Yes | Yes |
| FC | Yes | Yes | Yes | Yes | Yes | Yes | Yes | Yes | Yes |

*3.2.8. Training process*

The goal in the training process is to learn parameters of layers' filters in such a way that a loss function is minimized. We utilize the categorical cross-entropy as the loss function to compute the error in the output layer. We use the Adam optimizer to update model parameters in the back propagation process. Adam is a well-suited optimization technique for problems with large dataset and parameters that has recently seen broader adoption for deep learning applications (Kingma and Ba, 2014). We use the batch size equal to 64 and Adam's default settings as provided in their paper: learning rate = 0.001, $\beta_1 = 0.9$, $\beta_2 = 0.999$, and $\varepsilon = 10^{-8}$. We initialize the parameters in the convolutional and fully connected layers by following the proposed scheme in Glorot and Bengio (2010). In our initial analysis, we use a low number of epochs equal to 10 for identifying the optimal CNN configuration in Table 1 and the optimal combination of *P* values in dropout layers. Afterward, we apply the early stopping method to identify the optimal number of epochs for training the best identified CNN net, which avoids overfitting problem. In the early stopping method, the training and validation scores (e.g., accuracy) are computed after each epoch of training. The number of epochs that results in the maximum validation score is selected as the optimal value of epochs.

## 4. Data description and creation of GPS segments

The proposed methodology is examined and validated on the GPS trajectories collected by 69 users in the GeoLife project (Zheng et al., 2008b,a). Although many kinds of transport modes have been labeled by users, we only consider the ground transportation modes. In accordance with the user guideline linked to the published dataset (Zheng, 2012), we assign driving as the label of both taxi and car. Also, we refer all reported rail-based modes (e.g., subway, train, and railway) to train due to the layout of the rail-based system explained in the user guide. Therefore, our final list of transportations modes is: walk, bike, bus, driving, and train. After matching each user's label file with its corresponding GPS trajectories file, the user's GPS track is divided into trips if the time interval between two consecutive GPS points exceeds twenty minutes as the interval threshold (Zheng et al., 2008b). Next, trips are converted to into fixed-size segments when the number of GPS points for each segment is set to $M = 200$. The number 200 is the median of the number of GPS points in all trips. This is also used as the segmentation procedure for dividing the unseen GPS tracks before the classification task. So our proposed method is capable of classifying a GPS track that lasts approximately 10 min since more than 91.5% of trajectories has been recorded in a dense representation (e.g., every 1–5 s) (Zheng, 2012). The two consecutive segments with an identical label are then concatenated together. After calculating the motion characteristics of each GPS point, all the data pre-processing steps, described in Section 3.1, are applied to the created segments. The maximum allowable speed and acceleration pertaining to each mode are provided in Table 2. Using several reliable online sources and the engineering justification (e.g., existing speed limits, current vehicle and human's power), the speed and acceleration thresholds are designated so as to reject only the GPS points with unlikely speed and acceleration values. The segments with less than 10 GPS points are discarded. The window size and polynomial order in the Savitzky-Golay filter are set to 9 and 3, respectively. The distribution of the created samples among the modes are also illustrated in Table 2, with the total number of segments equal to 32,444.

Table 2
Number of samples, maximum speed, and maximum acceleration associated with each transportation mode.

| Transportation mode | Number of segments | Maximum speed (m/s) | Maximum acceleration (m/s$^2$) |
|---|---|---|---|
| Walk | 10,372 | 7 | 3 |
| Bike | 5568 | 12 | 3 |
| Bus | 7292 | 34 | 2 |
| Driving | 4490 | 50 | 10 |
| Train | 4722 | 34 | 3 |

## 5. Result and discussion

All data processing has been coded in the Python programming language. The CNN architectures have been implemented in Keras (Chollet, 2015), a Python-based deep learning library, using the TensorFlow backend with the CPU support only. We randomly sample 80% of the whole created segments as the training set while holding out the rest as the testing data. The final performance evaluation of a CNN model needs to be done only on the test set that plays no role in the training process.

### 5.1. Identifying the optimal CNN configuration

In the first round of our experiment, we seek for an optimal CNN configuration in terms of the layers' pattern, the absence or presence of a layer, the CNN's depth, and the layers' number of filters. The test accuracy rates of the most important CNN arrangements in Table 1 are set out in Table 3.

In the first three configurations, we only use convolutional layers. Increasing the number of convolutional layers from 2 in the model A to 6 in the model C enhance the accuracy of model by nearly 2%. Note that we increase the number of filters while the CNN's depth rises so as to capture more abstract features in last layers before the classification task. Adding an FC layer in the model D cannot improve the performance by itself. In order to assess the effect of max-pooling layers, we add a max-pooling layer right after each group of convolutional layers in the network E, which increase the test accuracy a bit. Analogously, we insert dropout layers after max-pooling and FC layers to avoid potential overfitting in the network F. As can be seen in Table 3, the accuracy dramatically goes down to 69% from the previous best performance 77% in the model E. The rationale behind such a significant decrease in accuracy is that using many dropout layers simplifies model too much and in turn causes high bias. Accordingly, a proper arrangement of dropout layers can make a balance between bias and variance (i.e., overfitting and underfitting problems). So we remove the dropout layers in the first groups of convolutional layers, which results in increasing the accuracy to its highest value of 79.8% in the model G. Keep the layout of model G, we also create deeper and more complex CNNs in models H and I by adding more convolutional and FC layers. In spite of achieving nearly similar results, the test accuracy has not improved in comparison with the configuration G. Another advantage of the network G is its less computational cost due to being shallower than models H and I.

In the second round of our experiment, we attempt to ameliorate the best CNN configuration G by first optimizing the P values in the dropout layers and then identifying the proper number of epochs for training. We specify the range of P values for both dropout layers in the model G as [0.2, 0.3, 0.4, 0.5, 0.6, 0.7, 0.8]. By exhaustively considering all combinations of P values and using a 5-fold cross validation over the training set, we obtain the highest score when both P values are set to 0.5, the same as the original model G. To obtain the optimal number of epochs, we train the model G over 120 epochs while the model performance is computed on both training and test sets at the end of each epoch. The optimal number of epochs is the point when no further increase occurs on the test score. The method is also known as early stopping. Accordingly, Fig. 3 compares the test and training scores for varying number of epochs. As expected, the training accuracy goes up by increasing the number of epochs and peaks at almost 96%. However, the test accuracy as the main evaluation metric levels off around 81–82% after around 30 epochs of training. The highest test accuracy is 82.3%, achieved with 62 epochs of training. Since the test accuracy remains almost constant and does not drop with increasing the number of epochs, we ensure the overfitting problem does not occur in our CNN model. The main remedy for a machine learning algorithm that performs well on training set but not on test set is to increase the training data. In other words, lack of training data in our application is the principal reason for inability to enhance the test accuracy while the model performs well on the training set yet with no overfitting problem. Also, the training time is affordable. It takes about 25 minutes to train the model G for 62 epochs using a system equipped with a Core i7 2.50 GHz processor and a 16.0 GB memory. As a consequence, we select the single configuration G with its optimized P values, which has been trained by 62 epochs, as the excellent architecture for applying the ensemble concept. We train seven CNNs with the same pattern of the model G but on separate training sets that have randomly been sampled with replacement from the original training set. Our ensemble yields the highest accuracy of 84.8%, a 2.5% increase compared to the model G. The results of test accuracy for different models in the second round of our experiment are summarized in Table 4.

Table 3
Test accuracy rate of the CNN configurations in Table 1.

| CNN Configuration | A | B | C | D | E | F | G | H | I |
|---|---|---|---|---|---|---|---|---|---|
| Accuracy (%) | 74.5 | 75.0 | 76.9 | 75.1 | 77.0 | 69.2 | **79.8** | 79.6 | 78.3 |

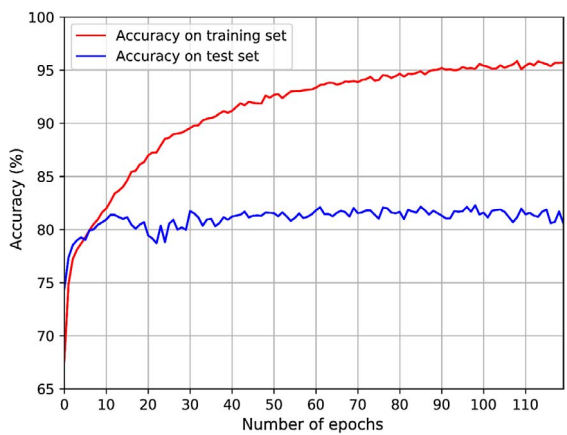

**Fig. 3.** Comparing the test and training accuracy for varying number of epochs on the best CNN configuration (model G).

**Table 4**
Test accuracy of various versions of the best CNN configuration (model G).

| Improvement steps for the best CNN net (model G) | Test accuracy (%) |
|---|---|
| Original model G in Table 1 | 79.8 |
| Model G with optimal P values of dropout layers | 79.8 |
| Model G trained with optimal number of epochs 62 | 82.3 |
| Ensemble of model G | 84.8 |

Table 5 provides the details of our best CNN performance (i.e., the ensemble of the model G) including the confusion matrix, recall, and precision pertaining to each transportation mode. Recall for each mode implies the accuracy of the predictor for only that mode. However, precision is the fraction of true instances among all predicted instances for each mode. In general, the results, particularly the recall rates, demonstrate the effectiveness of the CNN architecture in inferring the transportation modes, in which all precision and recall values exceed 67%. As reported in Table 5, there is a high correlation between the performance of the model in predicting a mode and the number of available instances for that mode. A large volume of samples and exclusive features of the walk mode results in achieving a perfect recall of 95.7%, whereas the driving mode with the lowest number of available segments obtains the lowest accuracy of 67.4%. It should be noted that the travelers' behavior in the driving mode is much more unpredictable since a wider range of taxi and passenger cars exist that have more flexibility in maneuvers compared to other vehicles. The bus and train are the transit modes that must adhere to pre-defined routes and schedule, which leads to more predictable mobility behavior. Another interesting yet reasonable finding is that a large portion of false negative instances for the driving mode is bus, 171 out of 902. This highlights that the CNN topology requires more instances to distinguish between the driving mode and the bus, as the most identical mode to the car and taxi.

### 5.2. Comparison with classical machine learning algorithms

In order to assess the performance of our best CNN model in comparison with classical machine learning algorithms, we choose widely used methods for learning transportation modes based on hand-crafted features, including K-Nearest Neighborhood (KNN), RBF-based Support Vector Machine (SVM), and Decision Tree (DT). We also compare our CNN with Random Forests (RF), as a representative of ensemble algorithms, and Multilayer Perceptron (MLP), as a regular and fully connected neural networks. All models are implemented and evaluated using scikit-learn, a Python-based machine learning library.

**Table 5**
Confusion matrix, recall, and precision for the ensemble of the model G.

| Ensemble of the model G | | Predicted class | | | | | | |
|---|---|---|---|---|---|---|---|---|
| | | Walk | Bike | Bus | Driving | Train | Sum | Recall% |
| Actual class | Walk | 2014 | 53 | 40 | 3 | 2 | 2112 | 95.7 |
| | Bike | 171 | 946 | 32 | 3 | 1 | 1153 | 82.6 |
| | Bus | 147 | 23 | 1104 | 70 | 25 | 1369 | 81.1 |
| | Driving | 71 | 17 | 171 | 601 | 42 | 902 | 67.4 |
| | Train | 75 | 17 | 31 | 24 | 806 | 953 | 85.3 |
| | Sum | 2478 | 1056 | 1378 | 701 | 866 | 6489 | – |
| | Precision% | 81.6 | 90.3 | 80.7 | 86.6 | 92.3 | – | – |

Table 6
Comparison of CNN's inference accuracy with five classical machine learning algorithms. HP: Hyperparameter, NA: Not applicable.

| Model | HP name | HP range | Optimal HP value | Test accuracy (%) | Average precision (%) | Average recall (%) | Average F-score (%) |
|---|---|---|---|---|---|---|---|
| KNN | No. of neighbors | [3, 40] | 5 | 63.5 | 63.2 | 63.5 | 62.4 |
| SVM | Regularization | [0.5, 20] | 4 | 65.4 | 66.9 | 65.4 | 64.9 |
| DT | Max. depth of tree | [1, 40] | 10 | 75.2 | 75.2 | 75.2 | 74.9 |
| RT | No. of trees | [5, 100] | 85 | 78.1 | 78.2 | 78.1 | 77.7 |
| MLP | No. of hidden layers | [1, 10] | 1 | 59.4 | 58.8 | 59.4 | 54.6 |
| CNN-A | Already satisfied | NA | NA | 75.6 | 75.2 | 75.6 | 74.8 |
| Best CNN | Already satisfied | NA | NA | 84.8 | 86.3 | 82.4 | 83.9 |

To have a fair comparison, the traditional inferring algorithms are trained and tested using the same training and test trajectory segments as were utilized in the CNN model. However, the input features for each segment needs to be manually designed. We utilize the same features as introduced in Zheng et al. (2008b,a), including the segment's length, mean speed, expectation of speed, variance of speed, top three speeds, top three accelerations, heading change rate, stop rate, and speed change rate. These features are the most acceptable GPS trajectories' attributes that have been utilized by many researchers.

Using the grid search method and 5-fold cross validation on the training set, the estimators' hyperparameters are optimized to ensure a good fit of the data and have the best trained model. The most important hyperparameters are enumerated as the number of neighbors for KNN, the maximum depth of tree for DT, the regularization parameter for SVM, the number of weak learners (i.e., trees) for RF, and the number of hidden layers for MLP. In the MLP model, we set the number of neurons in the preceding hidden layer twice as the number of neurons in its previous layer. After tuning the hyperparameters through an exhaustive search, the optimal estimators are evaluated on the test set. Table 6 compares the performance quality of our best CNN model with the above listed supervised learning algorithms on the similar test data in terms of four classification metrics: test accuracy, average precision, average recall, and average F-score. It should be noted that the reported precision, recall, and F-score are the average among all modes of transportation. The optimal hyperparameter of each estimator over the specified parameter range is also shown in Table 6. The comparison proves the superiority of our CNN model, in which the test accuracy of our best CNN is almost 16% higher than the average test accuracy of other methods. Almost the same results have been obtained for other performance measures. We also seek to compare the prediction quality of the shallowest and simplest CNN configuration in Table 1 (i.e., CNN-A) with other classical techniques. With regard to Table 6, it is interesting that the tuned CNN-A with only two convolutional layers still outperforms KNN, SVM, DT, and MLP models. The test accuracy of the RF model, as the best classic model, is better than the shallowest CNN model by only 2.5%. Another interesting finding is that our best CNN and the shallowest CNN significantly outperform the regular neural network (i.e., MLP). As was previously mentioned, the power of the CNN algorithm in extracting higher-level features through multiple layers of nonlinear processing units results in outperforming the traditional supervised learning algorithms that are fitted based on feature engineering.

### 5.3. Comparison with the previous studies

To show the superiority of our model, we compare our results with the studies that have used the GeoLife GPS trajectory dataset in order to establish a fair comparison. We choose the seminal study of Zheng et al. (2008a), who developed a solid inference mode framework for the first time in 2008, and the studies that employed the deep learning approaches to extract high-level features (Endo et al., 2016; Hao et al., 2017). Table 7 compares the prediction performance of our best CNN model with the highest performance accuracy reported by these studies. As can be seen in Table 7, our best CNN architecture significantly outperforms the frameworks in the mentioned studies by improving the accuracy more than 8%, 16%, and 10% compared to the work of Zheng et al., Endo et al., and Wang et al., respectively. This cutting-edge accuracy stems from our data pre-processing steps, data augmentation, regularization methods, an effective layout of the input layer, involving the fundamental motion characteristics, and of course the inherent ability of the CNN scheme in extracting high-level features. We conclude the shortage of a massive GPS dataset acts as the major deterrent for further progress in our CNN framework.

Table 7
Performance comparison with the seminal and most related studies.

| Model | Test accuracy (%) |
|---|---|
| Seminal study: Zheng et al. | 76.2 |
| Related study: Endo et al. | 67.9 |
| Related study: Wang et al. | 74.1 |
| Best model in this study: Ensemble of the model G | 84.8 |

## 6. Conclusion

In this article, we deployed the CNN architectures to infer transportation modes using only raw GPS trajectories. We established an unprecedented and state-of-the-art layout for the input layer of CNN architectures. We create a four-channel input volume, in which each channel represents a sequence of the basic kinematic attributes including speed, acceleration/deceleration, jerk, and bearing rate. We applied several data preprocessing steps to eliminate anomalous GPS logs and random errors. Moreover, we divided the long segments into fixed length samples, which is a requirement in the CNN method. Since a good performance of the CNN is primarily contingent on access to the high number of samples, the subdivision process was very effective in enhancing the accuracy of the model by augmenting the number of samples more than four times. Afterward, we examined a set of CNN architectures with different layers' patterns so as to identify the most efficient one. We improved the prediction performance of the optimal CNN configuration by increasing the number of epochs in the training process. Using the early stopping method, the optimal number of epochs was selected so as to avoid overfitting problem. The reported metrics indicate that our ensemble of seven best CNNs not only performs well on its own way but significantly outperforms the classical machine learning algorithms, including KNN, SVM, DT, RF, and MLP. Furthermore, the test accuracy of our best CNN model exceeds the highest test accuracy reported by previous studies. It is worth noticing that removing anomalies, designing an efficient input layer with the appropriate motion characteristics, augmenting training data, tuning some hyperparameters, and employing the bagging concept are the key factors in attaining such high accuracy. According to our observation and analysis, the lack of access to a larger GPS-trajectory dataset was recognized as the leading reason for not achieving further improvement in our proposed CNN framework. As a future research direction, using the available unlabeled GPS trajectories and leveraging the semi-supervised and unsupervised learning algorithms can be a potential compensation for not accessing to a large labeled dataset.

**Appendix A. Supplementary material**

Supplementary data associated with this article can be found, in the online version, at http://dx.doi.org/10.1016/j.trc.2017.11.021.

**Data Reference**